\pdfoutput=1

\documentclass[11pt]{article}

\usepackage[final]{acl}

\usepackage{times}
\usepackage{latexsym}
\usepackage{dingbat}
\usepackage[T1]{fontenc}
\usepackage[utf8]{inputenc}
\usepackage{microtype}
\usepackage{inconsolata}
\usepackage{booktabs}
\usepackage{multirow}
\usepackage{graphicx}
\usepackage{enumitem}
\usepackage{subfigure}

\title{Harnessing the Power of Multiple Minds: \\ Lessons Learned from LLM Routing}

\author{
    KV Aditya Srivatsa\thanks{Equal contribution} \\
    \And
    Kaushal Kumar Maurya\footnotemark[1] \\
    Mohamed bin Zayed University of Artificial Intelligence, Abu Dhabi, UAE \\
    \texttt{\{vaibhav.kuchibhotla, kaushal.maurya, ekaterina.kochmar\}@mbzuai.ac.ae} \\
    \And
    Ekaterina Kochmar
}

\begin{document}
\maketitle
\begin{abstract}
With the rapid development of LLMs, it is natural to ask how to harness their capabilities efficiently. In this paper, we explore whether it is feasible to direct each input query to a single most suitable LLM.
To this end, we propose \textit{LLM routing} for challenging reasoning tasks. 
Our extensive experiments suggest that such routing shows promise but is not feasible in all scenarios, 
so more robust approaches should be investigated to fill this gap.\footnote{Our code and data are available at \url{https://github.com/kvadityasrivatsa/llm-routing}.}
\end{abstract}
 
\section{Introduction}
\label{sec:intro}

Large language models (LLMs) demonstrate remarkable capabilities in many natural language generation and understanding tasks \cite{bommasani2021opportunities, chang2023survey, minaee2024large}. At the same time, \citet{jiang-etal-2023-llm} show that no single open-source LLM outperforms all others across different benchmarks and datasets, as various LLMs may exhibit different domain expertise~\cite{open-llm-leaderboard}. Experiments towards predicting model behavior~\cite{rabinovich2023predicting,srivatsa2024naacl} also suggest that particular aspects of input prompts can affect different LLMs in different ways. 

It is, therefore, reasonable to investigate whether the capabilities of different LLMs can be harnessed to achieve better results more efficiently. Recent findings suggest that performance can be improved with ensembling \cite{wang2022self, wang2023fusing, li2024more} and collaborative frameworks \cite{wu2023autogen, li2023camel}. However, the research in this area is still in the early stages, with a number of open research questions. In this work, we propose \textit{LLM routing}, which investigates whether {\em directing an input prompt to the most suitable single LLM can lead to better performance than what can be achieved with individual LLMs while maintaining a reasonable (e.g., single LLM) latency}.

With the rise of larger and more capable models in NLP and the wider field of ML, the research on sparse expert models has also extended. This class of models includes mixture-of-experts ~\cite{jacobs1991adapt, collobert2002scaling, eigen2014learning}, switch-transformers ~\cite{fedus2022switch}, and routing networks ~\cite{rosenbaum2017routing}, among other models.\footnote{For more details on the related work, see Appendix \ref{sec:related}.}  
Approaches to building these sparse models vary along several dimensions: (i) how the optimal parameter subset(s) or model-pool candidates are identified for each input (e.g., feature-based or deep-encoder-based classification), (ii) whether the subset selection involves pre-training the candidate models or model layers (e.g., Mixtral ~\cite{jiang2024mixtral}), which can incur significant training compute and data costs, (iii) how many experts are selected for each input (e.g., HybridLLM ~\cite{ding2024hybrid} selects only the single best, whereas \citet{shazeer2017outrageously} selects the top-k), and (iv) whether the approach also aims to improve the response quality or overall performance beyond that of any single candidate model. In this context, our paper aims to build and analyze a sparse LLM routing model that selects the single best LLM (from a pool of at least two LLMs) for each input query. The proposed router only requires fine-tuning of a relatively small pre-trained Transformer encoder model on the data without the need for pre-training or fine-tuning the LLMs. 

Given that LLMs frequently face challenges with reasoning and planning tasks \cite{wei2022chain, kojima2022large}, we focus on two well-established reasoning task benchmarks. We empirically investigate the feasibility of building \textit{LLM routing} model capable of selecting the most suitable LLM for each input from a pool of diverse LLMs. The routing is grounded on responses generated by LLMs. We explore binary and multi-label classification modeling at the input query level, as well as a clustering approach based on the similarity between the queries. Finally, leveraging prediction confidence scores, we design multiple optimal policies to select a single suitable LLM from the pool.

The contributions and key findings of this work are as follows: (1) We propose an LLM routing model, which directs input queries to the most suitable {\em single} LLM. (2) We explore two different types of approaches for LLM routing, treating it as a classification and a clustering task. (3) We conduct experiments with 7 open-source LLMs and on two challenging reasoning benchmarks ({\tt GSM8K} and {\tt MMLU}). (4) We introduce theoretical upper bounds for two scenarios: (i) highest possible performance achieved jointly with all LLMs (i.e., oracle), and (ii) highest performance achieved with the proposed routing model. (5) Our findings indicate that theoretical upper bounds of the routing model are higher than individual model performance, however, the proposed model developed in practice is unable to achieve those scores. Specifically, the performance of the routing model is better than that of the weak LLMs but is similar to or slightly lower than that of the top-performing LLMs, which may be due to the small size of the training data.

Despite the somewhat negative results, we believe this study demonstrates the feasibility of modeling LLM routing and contributes to new research directions on efficient usage of LLMs, which can benefit researchers and practitioners. 

\begin{figure}[!t]
    \centering
    \includegraphics[width=1\linewidth]{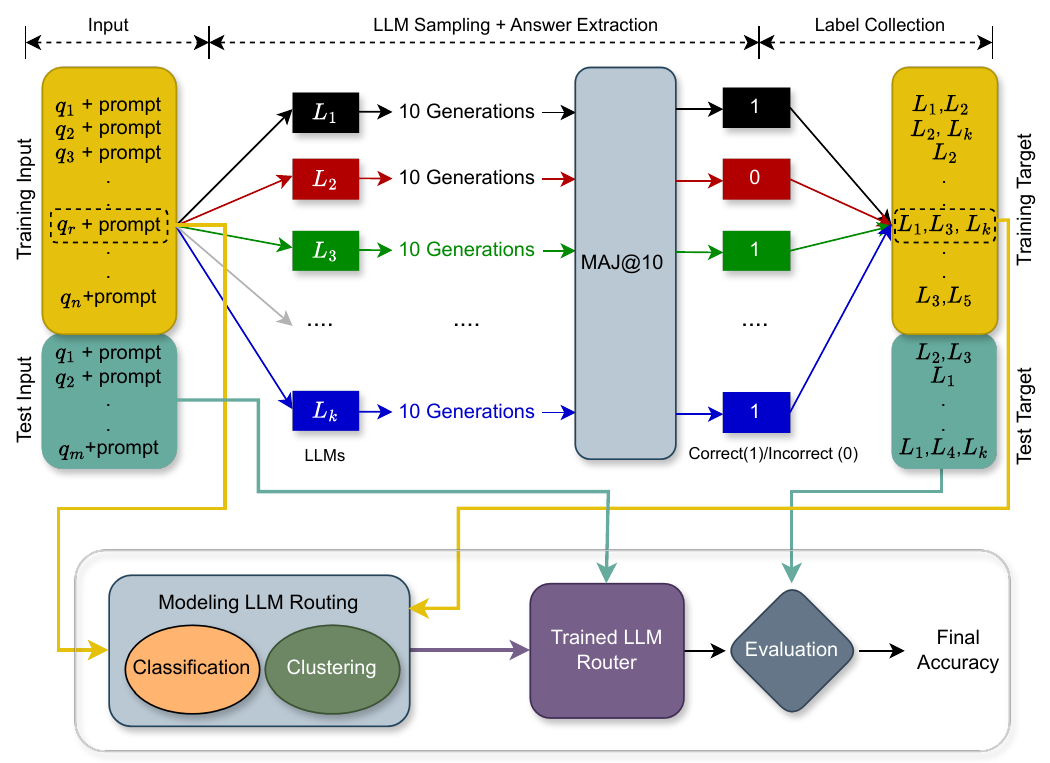}
    \caption{Overview of the proposed workflow.}
    \label{fig:workflow}
\end{figure}

\section{Methodology}

We present an overview of the proposed workflow in Figure \ref{fig:workflow}. Below, we describe our approaches to {\em LLM sampling} and {\em LLM routing}.

\subsection{LLM Sampling}

\paragraph{Selection of Benchmarks and LLMs}
As it has been observed that most of the existing LLMs struggle with reasoning tasks \cite{patel-etal-2021-nlp, wu-etal-2023-chain}, we focus on two challenging datasets associated with distinct domains -- mathematical (\texttt{GSM8K} by \citet{cobbe2021training}) and natural language reasoning (\texttt{MMLU} by \citet{hendryckstest2021}). {\tt GSM8K} consists of 8,792 diverse grade-school level math word problems (MWPs), while {\tt MMLU} contains 15k multiple-choice questions spanning 57 subjects across STEM, humanities, and social sciences, among others (see Table \ref{tab:data_stats}). We have selected diverse LLMs based on criteria such as performance on benchmarks, training methodologies, model specialization, and more. The final set of LLMs is presented in Table \ref{tab:llms}.

\begin{table}[!t]
    \centering
    \resizebox{0.9\columnwidth}{!}{%
    \begin{tabular}{l|r|r}
    \hline \hline
       \textbf{Split/Criteria}  & \textbf{GSM8K} & \textbf{MMLU} \\ \hline \hline
        Training & 6,816 & 13,757 \\ \hline
        Validation & 359 & 285 \\ \hline
        Test & 1,319 & 1,530 \\ \hline
        \#examples for few-shot CoT & 5 & 5 \\
        \hline \hline
    \end{tabular}%
    }
    \caption{Dataset statistics for the {\tt GSM8K} and {\tt MMLU} datasets. {\tt MMLU} data splits are remapped to have a distribution similar to {\tt GSM8K}. {\tt CoT}: Chain-of-Thought}
    \label{tab:data_stats}
\end{table}

\begin{table}[!t]
    \centering
    \resizebox{0.9\columnwidth}{!}{%
    \begin{tabular}{l|c|c|c}
    \hline \hline
       \textbf{LLMs}  & \textbf{Chat?} & \textbf{Specialized?} & \textbf{\#Parameters}\\ \hline \hline
        \texttt{llama2-7b} & $\times$ & $\times$ & 7B \\ \hline
        \texttt{llama2-13b-chat} & \checkmark & $\times$ &  13B \\ \hline
        \texttt{mistral-7b} &  $\times$ & $\times$ & 7B\\ \hline
        \texttt{mistral-7b-it} & \checkmark & $\times$ & 7B \\ \hline
        \texttt{gemma-7b} & $\times$ & $\times$ & 7B\\ \hline
        \texttt{gemma-7b-it} & \checkmark & $\times$ & 7B \\ \hline
        \texttt{metamath-7b} & $\times$ & \checkmark & 7B\\
        \hline \hline
    \end{tabular}%
    }
    \caption{List of diverse LLMs selected in this study.}
    \label{tab:llms}
\end{table}

\paragraph{Routing Data} In this study, we assess each LLM's performance by generating 10 responses for each input query to ensure more reliable and replicable behavior in our modeling. For LLM prompting and answer extraction from responses, we have followed the standard guidelines (see Appendices \ref{sec:pmtsamp} and \ref{sec:ansext} for details). Figure \ref{fig:samprmpt} presents the sample prompting templates. We use majority voting scores as labels for each input query to train routing classifiers. \textit{Majority Voting} ({\sc maj@k} $\in \{0,1\}$) determines whether the most frequent answer matches the gold answer or not. The mean {\sc maj@10} scores across all input queries are reported in Table \ref{tab:routing_merge}. Furthermore, to ensure a reliable response from an LLM, we consider only those LLMs for which the extracted answer viability scores are above 90\% (please refer to Appendix \ref{sec:pmtsamp} for more details), resulting in 6 viable LLMs for the {\tt GSM8K} dataset and 7 for the {\tt MMLU} dataset, respectively. We prepare the routing dataset by associating each input query with those viable LLM(s) that have a {\sc maj@10} score of 1. Formally, the target label for an input query $q \in Q$ is given by $label\left ( q \right ) = \left \{ l\: |\: l \in L, maj@10\left ( q,l \right ) = 1 \right \}$, where $L$ is the set of candidate LLMs and $Q$ is the set of query prompts from {\tt GSM8K} or {\tt MMLU}. 

\subsection{LLM Routing}
\label{ssec:routing}

Next, we build an LLM router, \textit{determining which model to select from a pool of LLMs for a given input query based on performance and inference latency.} The ideal routing algorithm should select an optimal single LLM with high accuracy and low latency. To this end, we explore modeling at the individual query level using classification and utilize similarities among queries using clustering.

\paragraph{Classifier-Based Routing}
The classification-based routing consists of (1) the development of a classifier that can predict a set of LLMs capable of solving the input query along with prediction confidence scores, and (2) the identification of the policy to select optimal LLMs (with high accuracy and low latency) from the predicted LLMs based on confidence scores in the range [0-1].

\noindent\textit{Multi-label and Separate Classifiers:} We have considered two types of classifiers: a multi-label classifier (\textsc{mlc}) and separate classifiers (\textsc{sc}). \textsc{mlc} aims to predict all LLMs apt for a given input query together in a single prediction step. The \textsc{sc} model, on the other hand, employs a separate binary classifier for each LLM and accumulates the results post hoc. Both types of classifiers are built on top of existing popular pre-trained language models (PLMs). Specifically, we experimented with \texttt{BERT}, \texttt{DistilBERT}, \texttt{RoBERTa}, and \texttt{T5} models. Additionally, due to the small size of the training data, we explored smaller models, utilizing only a few layers of PLMs, as well as simpler models such as Random Forests. \texttt{RoBERTa} emerged as the best-performing model, and all results in this paper are reported with classifiers built by fine-tuning the \texttt{RoBERTa} PLM exclusively.

\noindent\textit{Proposed Policies:}
We utilize the classifiers' predicted confidence scores to design the following policies:
\begin{enumerate}[itemsep=-0.3em]
    \setlength{\leftmargin}{0pt}
    \item \textbf{ArgMax:} Select an LLM with the highest confidence score.
    \item \textbf{Random:} Select a pool of LLMs with confidence above a certain threshold (i.e., 0.80) and randomly pick one LLM from the pool.
    \item \textbf{Prediction:} Train a \texttt{RandomForest} regressor using training data confidence scores, where each input represents the confidence score for each predicted label, and the target is the confidence score of the first gold reference LLM. At test time, we select the LLM with a confidence score closest to the predicted score.
    \item \textbf{Sorted Prediction (Sorted Pred):} Similarly to the `Prediction' policy, the input confidence scores are arranged in ascending order based on LLMs' performance. This ensures that weaker LLMs have a fair opportunity.
\end{enumerate}

\paragraph{Clustering-Based Routing}
Additionally, to incorporate the query-level similarities, we explore clustering for LLM routing as detailed below.

\noindent\textit{Learning Clusters:} We fit a KMeans\footnote{\url{https://scikit-learn.org/stable/modules/generated/sklearn.cluster.KMeans.html}} clustering model on query-specific features extracted from the training data to learn discrete clusters. The features are extracted using: (1) TF-IDF vectorizer,\footnote{\url{https://scikit-learn.org/stable/modules/generated/sklearn.feature_extraction.text.TfidfVectorizer.html}} and (2) pooled hidden embedding of the \texttt{RoBERTa}\footnote{\url{https://huggingface.co/FacebookAI/roberta-base}} model's last layer.

\noindent\textit{Routing:} For each cluster in the training set, the best performing LLM is determined. At inference, input queries in the test set are routed to the best-performing LLM for their corresponding cluster.

\section{Experimental Setup}
\paragraph{LLM Routing Baseline Models} The following baseline models are included for comparison:
\begin{enumerate}[itemsep=-0.3em]
    \setlength{\leftmargin}{0pt}
    \item \textbf{Oracle:} The maximum possible performance is assumed under the premise that an oracle always selects a single LLM capable of solving each query if it is solvable.
    \item \textbf{Random:} This represents the mean performance of randomly selecting an LLM uniformly for each input query across 1000 independent runs.
   \item \textbf{Individual Models:} This is the mean performance of individual models with {\sc maj@10} across all queries. 
   \item \textbf{All LLMs:} This baseline reports the mean accuracy of {\sc maj@}(10$\times |L|$) based on the combined pool of 10 generations from each LLM, where $|L|$ is the total number of LLMs.
\end{enumerate}

\paragraph{Classifier Upper Bound} This is similar to the oracle model, where the upper bound is calculated with predicted labels instead of gold labels. 

\section{Results and Discussion}

\begin{table}[!t]
\resizebox{\columnwidth}{!}{%
\begin{tabular}{ll|cc|cc}
\hline
\hline
\multicolumn{2}{l|}{\multirow{2}{*}{\textbf{Models}}} & \multicolumn{2}{c|}{\textbf{GSM8K}} & \multicolumn{2}{c}{\textbf{MMLU}} \\ \cline{3-6} 
\multicolumn{2}{l|}{} & {\sc \textbf{Acc}} & \textbf{{\sc \textbf{Lat}} (sec)} & {\sc \textbf{Acc}} & {\sc \textbf{Lat}} \textbf{(sec)} \\ \hline
\multicolumn{2}{l|}{Oracle} & 87.18 & 3.46 & 89.15 & 1.89 \\
\multicolumn{2}{l|}{Random} & 55.37 & 3.52 & 52.50 & 2.35 \\ \hline
\multicolumn{2}{l|}{\texttt{gemma-7b}} & \underline{71.11} & 7.10 & \underline{63.85} & 3.00 \\
\multicolumn{2}{l|}{\texttt{metamath-7b}} & 67.55 & 4.70 & 42.28 & 2.40 \\
\multicolumn{2}{l|}{\texttt{mistral-7b}} & 59.74 & 3.70 & 62.09 & 1.80 \\
\multicolumn{2}{l|}{\texttt{*mistral-7b-it}} & 50.41 & 1.00 & 51.63 & 1.10 \\
\multicolumn{2}{l|}{\texttt{llama2-13b-chat}} & 46.70 & 1.80 & 50.52 & 4.80 \\
\multicolumn{2}{l|}{\texttt{*gemma-7b-it}} & 36.84 & 0.70 & 49.28 & 1.00 \\
\multicolumn{2}{l|}{\texttt{llama2-7b}} & -- & -- & 48.36 & 2.30 \\ \hline
\multicolumn{2}{l|}{All LLMs} & 74.37 & 19.00 & 60.39 & 16.40 \\ \hline
\multirow{5}{*}{MLC} & Upper bound & 79.68 & 5.16 & 77.18 & 1.94 \\
 & ArgMax policy & 67.62 & 4.76 & 62.28 & 2.95 \\
 & Random policy & 67.47 & 4.76 & 58.16 & 2.86 \\
 & Prediction policy & \textbf{67.70} & 4.77 & \textbf{63.85} & 2.95 \\
 & Sorted Pred policy & 59.90 & 4.77 & 48.36 & 2.92 \\ \hline
SC & ArgMax policy & 67.55 & 4.70 & 62.87 & 2.94 \\ \hline
\multirow{2}{*}{Clustering} & TF-IDF & 67.55 & 4.70 & 61.76 & 2.83 \\
 & RoBERTa & 67.55 & 4.70 & 61.76 & 2.83 \\ \hline
 \hline
\end{tabular}%
}
\caption{Performance of different routing models on {\tt GSM8K} and {\tt MMLU} test sets. For all queries, we have considered 10 generations with each LLM. {\sc Acc}: mean accuracy with {\sc maj@10} (\%), {\sc Lat}: LLM inference latency in seconds per query (10 generations for each query), {\sc mlc}: multi-label classifier, and {\sc sc}: separate classifiers. * The term `\texttt{it}' indicates instruction-tuned LLMs. The highest individual-LLM accuracy is \underline{underlined}, and the highest classifier accuracy is in \textbf{bold} for each dataset.}
\label{tab:routing_merge}
\end{table}

In Table \ref{tab:routing_merge}, we present the performance of each individual LLM across both datasets, alongside the performance of baselines and routing models. 
We observe that, even though {\tt gemma-7b} outperforms other LLMs, there are diverse performance trends for other LLMs across datasets, with some performing better on {\tt GSM8K}, and others on {\tt MMLU}. 
To investigate the results further, we pose and address a number of research questions.

\paragraph{Does including multiple LLMs solve all questions in a given dataset?} The Oracle model's {\sc Acc} scores for both datasets are lower than 90\%, indicating that more than 10\% of questions cannot be solved by all LLMs combined. For details, see Figure ~\ref{fig:sample_question} in the Appendix, where we project the distribution of questions solved by each of the LLMs.

\paragraph{How effective is a routing model when randomly picking LLMs?} As expected, the random baseline model achieves the lowest  {\sc Acc} score for both datasets. This highlights the necessity for an effective routing model to navigate through LLMs.

\paragraph{Is the joint performance of multiple LLMs better than that of individual LLMs?} Considering extreme cases like top-$k$ and bottom-$k$ LLMs as shown in Appendix Tables \ref{tab:det_routing_gsm8k} and \ref{tab:det_routing_mmlu}, we find that multiple LLMs collectively outperform single LLMs in terms of  {\sc Acc}. Even the joint performance with the bottom-2 model is better than that of individual models, underscoring LLMs' diverse problem-solving capabilities. However, we note two limitations in joint modeling: (i) the joint performance with all LLMs may not always be the best (see \textit{All LLMs} baseline {\sc Acc} scores), as reported for the {\tt MMLU} dataset, and (ii) joint modeling drastically increases inference latency costs (i.e., {\sc Lat}), aligning with recent research \cite{li2024more}. In contrast, the proposed LLM routing aims to leverage joint LLM capabilities while minimizing latency by selecting the single best-suited LLM. 

\paragraph{Can the upper bound performance of the classifier/clustering be equal to the Oracle model performance?} This is possible in an ideal scenario where classifier/clustering routing algorithms are perfect and bias-free. However, in our case, the training data for the algorithms is small ($\sim$9k in {\tt GSM8K} and 15k in {\tt MMLU}), which leads to sub-optimal performance. Still, the multi-label classifier's upper bound ({\sc Acc}) has achieved a higher score than any individual LLMs, which is also close to the Oracle model performance. We hypothesize that more training data for classification/clustering may bridge this gap.

\paragraph{Does router modeling with multi-label classifiers exhibit better performance than individual LLMs?} Unfortunately, the proposed multi-label classifier with different confidence-based policies does not lead to better performance (i.e., {\sc Acc}) than some individual LLMs. This may be due to the small training data for the classifier. However, it can be observed that the classifier's performance is better than most of the weak-performing LLMs and close to the top-performing LLM. This suggests that LLM routing is a promising direction that requires better classifier modeling.

\paragraph{What is the impact of different policies on LLM router modeling?} We have proposed four policies based on the label confidence scores of the multi-label classifier. The best policy can push the model performance closer to the upper bound performance of the multi-label classifier. However, we observe that due to the imperfect classifier (which yields weighted F1 scores of 0.71 for {\tt GSM8K} and 0.67 for {\tt MMLU}), the predictions (and confidence scores) are skewed towards only a few labels (see Figure \ref{fig:llmsdist} in the Appendix) which leads to sub-optimal  {\sc Acc} score. The predictions-based policy is better than other policies; however, the classifier performance presents a serious bottleneck. We conclude that larger training data and the development of a better classifier are essential for improving the  {\sc Acc} scores. Small sizes of both {\tt GSM8K} and {\tt MMLU} datasets prevent further investigation of this question.

\paragraph{How does a separate classifier compare to a multi-label classifier for LLM routing?} With relatively small and imbalanced training sets, separate classifiers for each LLM are more prone to over-fitting. Despite attempts to address this with measures like early stopping and weighted class-based loss, most individual models usually converge to the overall best performers such as \texttt{gemma-7b} on test split. Ultimately, with the argmax policy in place, the separate classifier-based routing model's performance is similar to that of the argmax policy of the multi-label classifier.

\paragraph{How does clustering-based LLM routing compare to other models?} The cluster-level routing approach aims to select the best LLM for a group of similar query prompts. It assumes that the relative performance of LLMs for each cluster remains consistent between the training and test sets. We find that this assumption does not hold for many clusters (39 out of 50 for {\tt GSM8K} and 28 out of 50 for {\tt MMLU}). In general, the best-performing LLM for most clusters in the training set is the same as the best LLM overall. The impact of different feature extraction methods (TF-IDF vs. {\tt RoBERTa}) was minimal, resulting in a similar performance to the {\sc mlc}+ArgMax model.

\paragraph{What is the impact of LLM routing on inference latency?} Table \ref{tab:routing_merge} provides the inference latencies for all LLMs, baselines, and LLM routing models in seconds per query, recorded using a single Nvidia A100 GPU. Ideally, the best routing policies should maximize model accuracy (while maintaining at least same-level latency) or minimize overall latency (with the best LLM accuracy maintained). For instance, the {\sc mlc}+ArgMax latency is lower than the corresponding highest individual model latency (of \texttt{gemma-7b}) for \texttt{GSM8K}. However, as the routing classifiers overfit to the best LLMs on the training sets (\texttt{metamath-7b} for \texttt{GSM8K} and \texttt{gemma-7b} for \texttt{MMLU}), the overall latency, much like mean accuracy, differs very slightly from that of the best LLMs. These findings validate our claim that the proposed \textit{LLM routing} model consistently maintains a latency score equal to or lower than any individual LLM. 

\paragraph{Ablations with multi-label routing:} In appendix Figure ~\ref{fig:ablation}, we overview ablation tests for LLM routing using a multi-label classifier trained with best- and worst-performing LLMs across both datasets. Key insights include: (1) Increasing the number of top-performing LLMs improves oracle scores but has marginal effects on the classifier's upper bound or argmax policy. (2) Increasing the number of worse-performing LLMs results in higher scores across oracle, {\sc mlc}'s upper bound, and {\sc mlc}+ArgMax policy model, highlighting the effectiveness of LLM routing.

\section{Conclusions and Future Directions}
This study investigates the feasibility of \textit{LLM routing}, i.e., navigating input queries by efficiently selecting the most suitable single LLM from a pool of LLMs. Through extensive experimentation with multi-label and separate classifiers, as well as clustering across two challenging benchmarks, we conclude that (i) there are theoretical bounds that can be achieved with LLM routing that are much higher than individual models' performance, and (ii) routing LLMs is a feasible direction that works best with equally capable LLMs. However, if a few LLMs dominate, the router's performance degrades, even though it still outperforms weak LLMs. At the same time, the inference latency of the routing model is at least at the same level as that of single LLMs.

With these findings in mind, we envision future research to investigate the following directions: (1) collecting larger datasets for LLM routing design; (2) developing novel models for LLM routing to accommodate LLMs with diverse capabilities; (3) designing better routing policies with confidence scores; (4) incorporating LLM-specific features for improved modeling; and (5) scaling up using more diverse LLMs and benchmarks.

\section*{Limitations} One of the key limitations of the proposed routing model is the limited training data available for training different algorithms with varying policies, which can result in biased learning despite taking a number of precautionary measures. Another limitation is the extraction of answers from generated responses: despite utilizing our best answer extraction algorithm, we could only extract viable answers for 83\% to 95\% of queries (with different LLMs). For the remaining queries, the answers extracted with our algorithm may be invalid or incorrect. Next, the proposed model works well with equally capable LLMs but is not yet effective enough for LLMs that have very different capabilities. 

Finally, although the inference latency of the proposed model is comparable to that of the most suitable single LLM, frequent switches between the LLMs (based on the input queries) necessitate loading most of the LLMs into memory, posing a limited memory issue. This issue is also observed with different emerging LLMs \cite{jiang-etal-2023-llm, jiang2024mixtral} similarly to our case. At the same time, the problem of limited memory in the context of LLMs has been well studied \cite{alizadeh2023llm, eliseev2023fast}, and the solutions developed are directly (or with minor adjustments) applicable to our modeling, thereby ensuring the practical usability of the proposed model. We leave investigation of such approaches to future work.

\section*{Ethics Statement}
This paper introduces router modeling to effectively harness the power of LLMs with different capabilities. As the proposed routing models use LLMs, we must acknowledge that, independently of this research, there are certain risks that pertain to all LLMs, as such models may generate outputs that, although plausible, are factually incorrect or nonsensical. Such \textit{hallucinations} can misguide decision-making and propagate biases, especially in critical scenarios where accuracy is vital. Without proper safeguards, widespread LLM adoption could worsen these concerns. Thus, it is essential to develop mechanisms to mitigate hallucination risks, ensuring responsible and beneficial deployment of these powerful models before adopting the proposed routing model.

\section*{Acknowledgments}
We thank Jad Doughman and Ted Briscoe for insightful discussions about this research. We are grateful to the Campus Super Computing Center at MBZUAI for supporting this work. We also thank the anonymous reviewers for their valuable feedback.

\bibliography{custom} 

\appendix

\newpage

\section{LLM Inference Latency}
\label{sec:llm_lat}

\begin{table}[!htb]
\centering
\resizebox{0.9\linewidth}{!}{%
\begin{tabular}{l|l|c|c}
\hline 
\hline
\textbf{Prompt Type} &  \textbf{LLM} & \textbf{GSM8K} & \textbf{MMLU} \\ \hline
& \texttt{llama2-7b} & 4.21 & 2.30 \\
& \texttt{gemma-7b} & 7.10 & 3.00 \\
FCoT & \texttt{mistral-7b} & 3.70 & 1.80 \\
& \texttt{metamath-7b} & 4.70 & 2.40 \\ \hline
& \texttt{gemma-7b-it} & 0.70 & 1.00 \\
ZCoT & \texttt{llama2-13b-chat} & 1.80 & 4.80 \\
& \texttt{mistral-7b-it} & 1.00 & 1.10 \\ \hline
\hline
\end{tabular}%
}
\caption{Statistics on the inference latency (i.e., runtime in seconds) for various LLMs over 10 generations for each input query. The timings were recorded using a single Nvidia A100 GPU. {\tt FCoT} denotes few-shot Chain-of-Thought, and {\tt ZCoT} denotes zero-shot CoT. We have considered 5 examples for {\tt FCoT} prompting.}
\label{tab:llm_lat}
\end{table}

\section{Prompting for LLM Sampling}
\label{sec:pmtsamp}

The consideration of diverse LLMs and datasets contributed to the challenges in prompting, as there is no single uniform prompting approach across LLMs and datasets \cite{sclar2023quantifying}. Considering recent findings about the appropriate usage of prompts \cite{sahoo2024systematic} and those from our own experimentation, we have converged on the following prompting decisions:

\begin{itemize}
\item For non-chat LLMs, few-shot Chain-of-Thought (CoT; \citet{wei2022chain}) prompting works better than zero-shot \cite{kojima2022large} for both datasets. We used 5 few-shot examples. The few-shot prompting leads to over 95\% \textit{viable} answers (except for {\tt llama2-7b} LLM, which has the viability score of 83\%) in generated solutions. A \textit{viable} answer is a single numeric/alphabetic answer that can be extracted from the generated solution using extraction algorithms (see Appendix \ref{sec:ansext}) to compare with the reference answer. The viable answer can be correct or incorrect.
\item For chat LLMs, few-shot CoT distracts the generation, which leads to unexpected outputs. The zero-shot CoT works best. We utilize different models' chat-templates from Hugging Face\footnote{\url{https://huggingface.co/docs/transformers/en/chat_templating}} to ensure correctness. The viability of answer extraction for chat models is 92\%.
\end{itemize}

\begin{figure*}[!htb]
    \centering
    \includegraphics[width=1\linewidth]{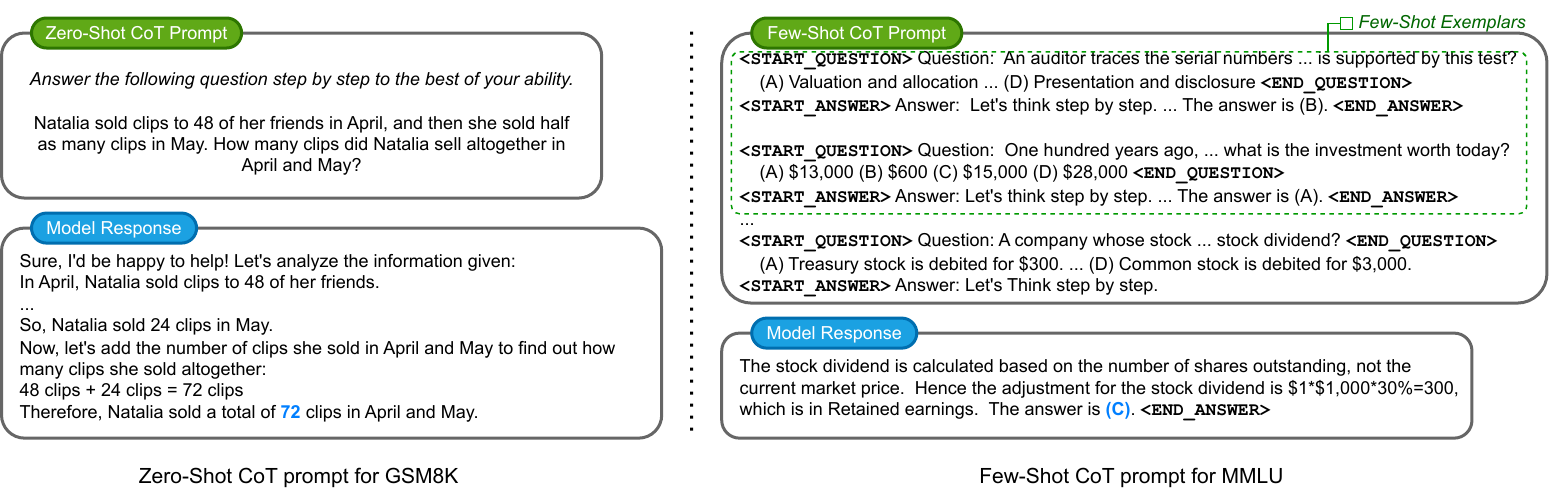}
    \caption{Sample zero-shot Chain-of-Thought (CoT) prompt template for a chat (or instruction-tuned) LLM and few-shot Chain-of-Thought (CoT) prompt template for a standard LLM.}
    \label{fig:samprmpt}
\end{figure*}

The sample zero-shot and few-shot CoT prompt templates are presented in Figure \ref{fig:samprmpt}.

\section{Answer Extraction from LLM Responses}
\label{sec:ansext}
The adapted prompting approaches used in our LLM queries are designed to instruct LLMs to specify their final answers at the very end of each of their responses. We thus use a simple answer extraction policy of selecting the last mentioned numerical value (for {\tt GSM8K}) and multiple-choice option (for {\tt MMLU}) from the generated responses. Responses failing to report any final answer are regarded as invalid and counted as incorrect answers. For {\tt MMLU}, we evaluate the extracted options directly against the annotated correct options (among `A', `B', `C', and `D') in the dataset. For {\tt GSM8K}, questions where the absolute difference between the ground truth and predicted numerical answers are less than $\epsilon=0.1$ are considered to be solved correctly. This threshold was set to accommodate instances where model-generated real-valued answers differ slightly from the expected answer.

\textbf{Lessons Learned:} It is observed that sometimes the expected answer is present in one of the last sentences of the response instead of at the very end. We extracted all such answers as well. Allowing a $0.1$ absolute error difference leads to more accurate answers.

\section{Implementation, Hyperparameters, and Hardware Details}
\label{sec:hyperparams_etc}

\paragraph{Querying LLMs} We use the vLLM package\footnote{\url{https://github.com/vllm-project/vllm}} to query LLMs. All models were queried with a temperature of 0.8 and a max token length of 2000. Each question prompt was queried 10 times with different initialization seeds. We used a single Nvidia A100 GPU for all runs. Querying each dataset once took approximately 1-2 hours.

\paragraph{Training Routing Classifiers} We use the HuggingFace library\footnote{\url{https://huggingface.co/}} for loading and tuning all pre-trained Transformer encoders in our experiments. Each model was trained for 10 epochs, with an initial learning rate of 2e-5, warmup ratio of 0.1, and class-balanced CrossEntropy loss. The training checkpoint with the lowest validation loss was selected for inference.

\label{sec:samprmpt}

\begin{table}[!htb]
    \centering
    \resizebox{0.8\columnwidth}{!}{%
    \begin{tabular}{l|c|c}
    \hline\hline
       \textbf{Models}  & {\sc \textbf{Acc} (\%)} & {\sc \textbf{Lat}} (sec)\\ \hline \hline
        Oracle & 87.18 & 3.46\\ 
        Random & 55.37 & 3.52\\ \hline
        \texttt{gemma-7b} & 71.11 & 7.10\\
        \texttt{metamath-7b} & 67.55 & 4.70\\ 
        \texttt{mistral-7b} & 59.74 & 3.70\\
        \texttt{mistral-7b-it} & 50.41 & 1.00\\
        \texttt{llama2-13b-chat} & 46.70 & 1.80\\  
        \texttt{gemma-7b-it} & 36.84 & 0.70\\ \hline 
        top-2 LLMs & 81.80  & 11.80\\ 
        top-3 LLMs & 84.00 & 15.5\\ 
        top-4 LLMs & 85.82 & 16.5\\ 
        top-5 LLMs & 86.03 & 18.3\\ \hline
        bottom-2 LLMs & 55.64 & 2.50\\ 
        bottom-3 LLMs & 67.02 & 3.50\\ 
        bottom-4 LLMs & 75.51 & 7.20\\ 
        bottom-5 LLMs & 79.91 & 11.90\\ \hline
        All LLMs & 74.37  & 19.00\\ \hline
        Upper Bound of MLC & 79.68 & 5.16\\ 
        MLC + Argmax policy & 67.62 & 4.76\\ 
        MLC + Random policy & 67.47 & 4.76\\ 
        MLC + Prediction policy & 67.70 & 4.77\\ 
        MLC + Sorted Pred policy & 59.90 & 4.77\\ \hline
        SC + Argmax policy & 67.55 & 4.70\\ \hline
        Clustering + TF-IDF & 67.55 & 4.70\\ \hline 
        Clustering + RoBERTa & 67.55 & 4.70\\ \hline 
        \hline
    \end{tabular}%
    }
    \caption{Performance of different routing models on {\tt GSM8K} data. {\sc Acc}: mean accuracy with {\sc maj@10} (\%), {\sc Lat}: LLM inference latency in seconds per query (10 generations for each query), {\sc mlc}: multi-label classifier, {\sc sc}: separate classifiers, and top-$k$: best $k$ performing models. All other notation is the same as for Table \ref{tab:routing_merge}.}
    \label{tab:det_routing_gsm8k}
\end{table}

\section{Detailed Results for Routing Models}

\begin{figure}[h!]
     \centering
     \includegraphics[width=0.95\columnwidth]{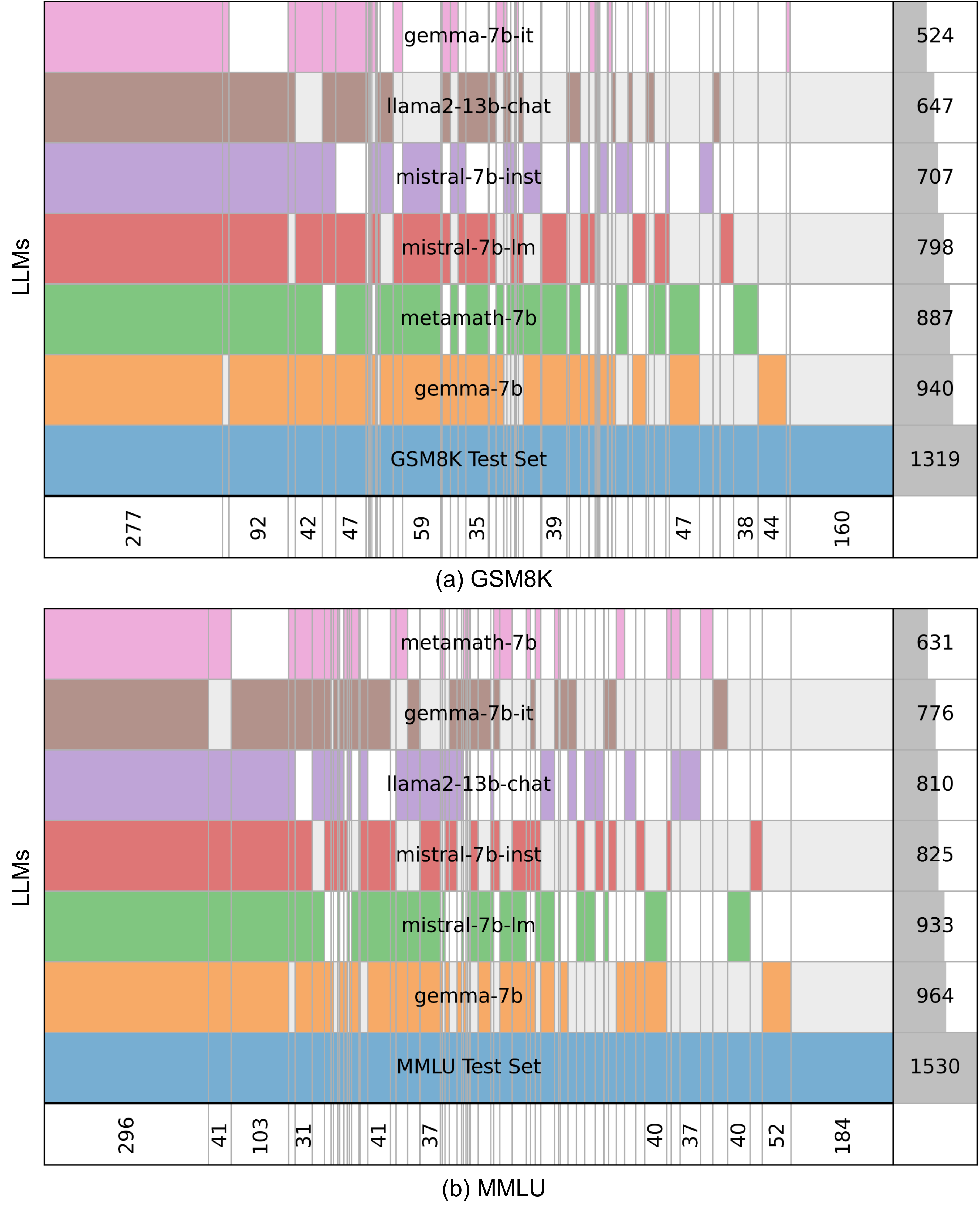}
     \caption{Distribution of queries from the {\tt GSM8K} and {\tt MMLU} test sets solved (score $1.0$ with {\sc maj@10}) by each LLM. The counts at the bottom of each figure denote the number of questions in each chunk, and those on the right denote the total number of questions solved by each LLM.}
     \label{fig:sample_question}
\end{figure}

\begin{figure}[!h]
    \centering
    \subfigure{\includegraphics[width=0.41\textwidth]{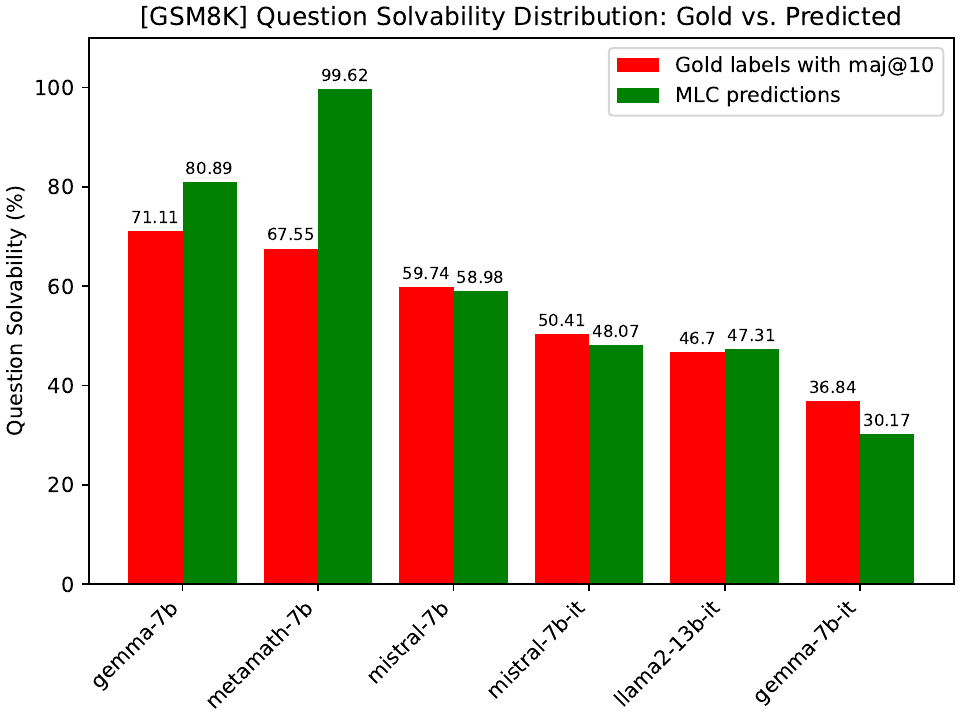}} 
    \subfigure{\includegraphics[width=0.41\textwidth]{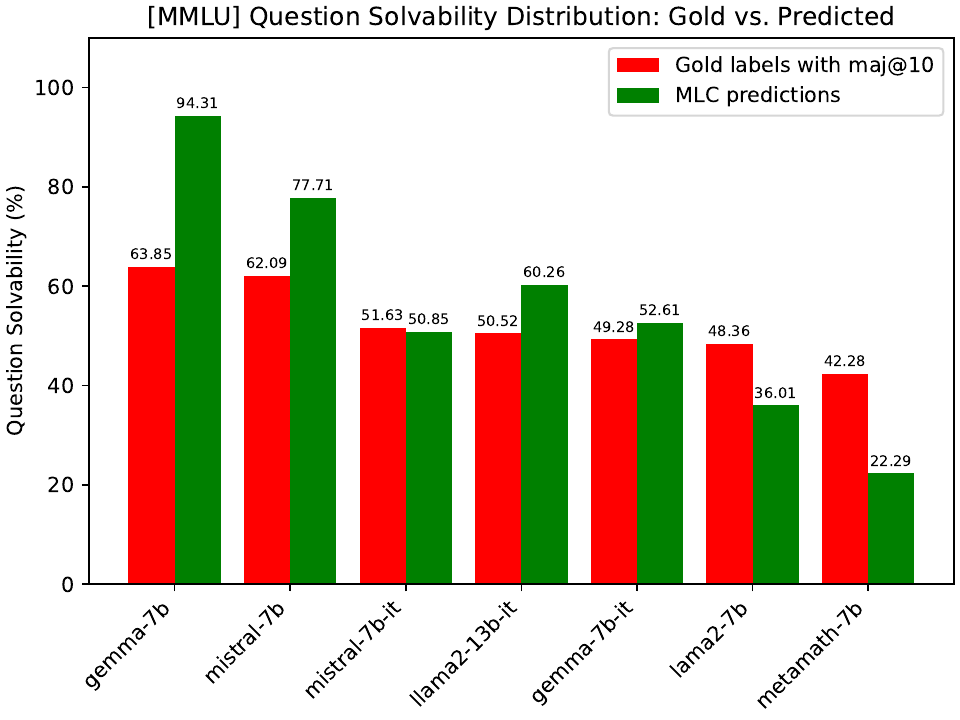}} 
    \caption{LLMs ``solvability" distribution. The gold label scores are obtained with {\sc maj}@10, and prediction label scores are obtained with a multi-label classifier.}
    \label{fig:llmsdist}
\end{figure}

\begin{figure*}[!t]
    \centering
    \subfigure{\includegraphics[width=0.35\textwidth]{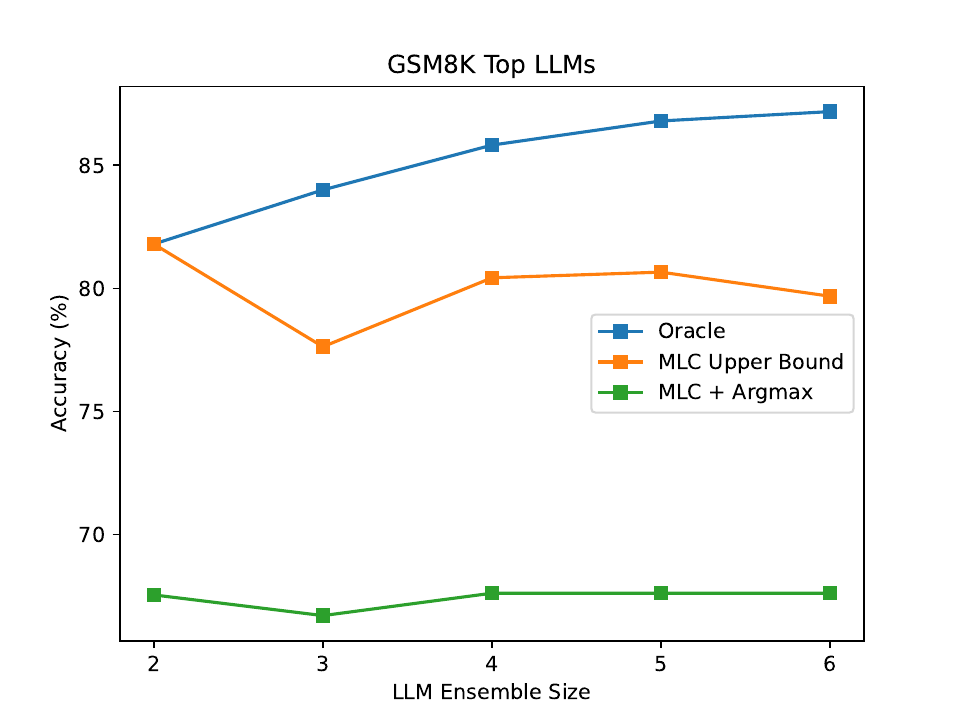}} 
    \subfigure{\includegraphics[width=0.35\textwidth]{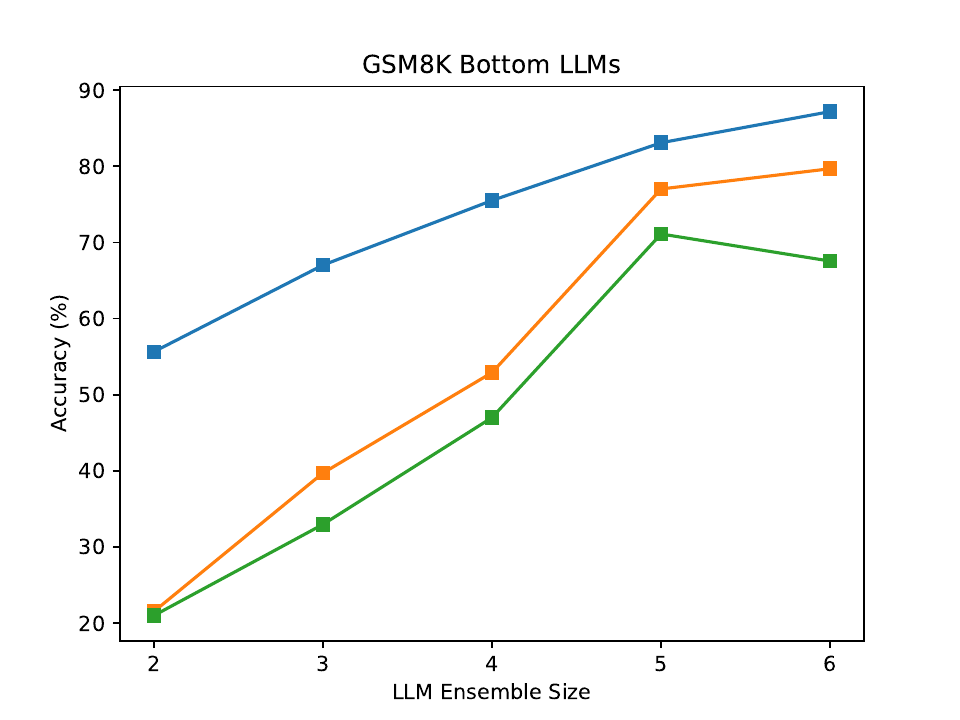}} 
     \subfigure{\includegraphics[width=0.35\textwidth]{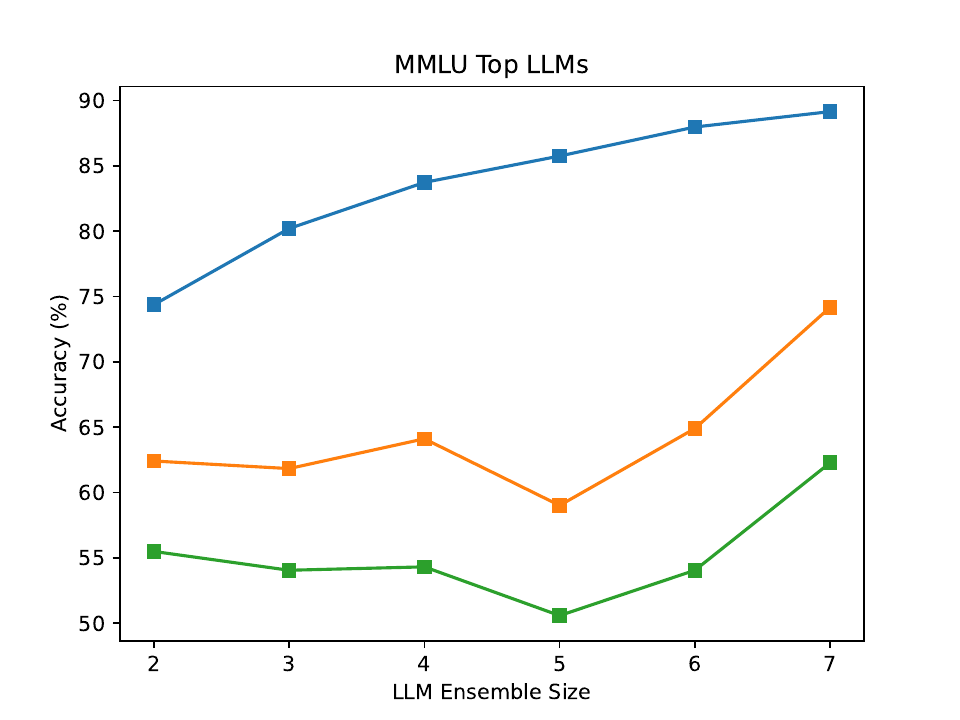}}
    \subfigure{\includegraphics[width=0.35\textwidth]{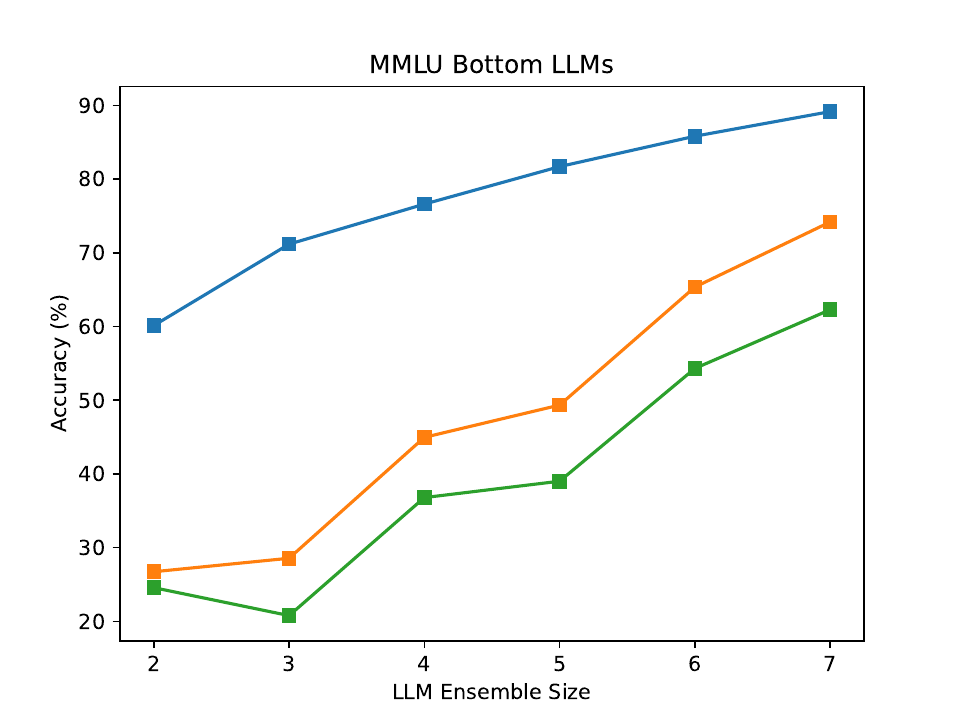}}
    \caption{Different ablation configurations 
    for LLMs for {\tt GSM8K} and {\tt MMLU} datasets.}
    \vspace{-0.4cm}
    \label{fig:ablation}
\end{figure*}

See Figures \ref{fig:sample_question}-\ref{fig:ablation} and Tables \ref{tab:det_routing_gsm8k}-\ref{tab:det_routing_mmlu}.

\begin{table}[!htb]
    \centering
    \resizebox{0.8\columnwidth}{!}{%
    \begin{tabular}{l|c|c}
    \hline\hline
       \textbf{Models}  & {\sc \textbf{Acc} (\%)} & {\sc \textbf{Lat}} (sec)\\ \hline\hline
        Oracle & 89.15 & 1.89\\ 
        Random & 52.50 & 2.35\\ \hline
        \texttt{gemma-7b} & 63.85 & 3.00\\
        \texttt{mistral-7b} & 62.09 & 1.80\\
        \texttt{mistral-7b-it} & 51.63 & 1.10\\
        \texttt{llama2-13b-chat} & 50.52 & 4.80\\
        \texttt{gemma-7b-it} & 49.28 & 1.00\\
        \texttt{llama2-7b} & 48.36 & 2.30\\
        \texttt{metamath-7b} & 42.28 & 2.40\\ \hline 
        top-2 LLMs & 73.47  & 4.80\\ 
        top-3 LLMs & 79.54  & 5.90\\ 
        top-4 LLMs & 83.72  & 10.70\\ 
        top-5 LLMs & 85.75  & 11.70\\ 
        top-6 LLMs & 87.88  & 14.0\\ \hline
        bottom-2 LLMs & 60.13  & 4.70\\ 
        bottom-3 LLMs & 71.17  & 5.70\\ 
        bottom-4 LLMs & 78.10  & 10.50\\ 
        bottom-5 LLMs &  81.69 & 11.60\\ 
        bottom-6 LLMs & 83.11  & 13.40\\ \hline
        All LLMs & 60.39  & 16.40\\ \hline
        Upper Bound of MLC & 77.18 & 1.94\\ 
        MLC + Argmax policy & 62.28 & 2.95\\ 
        MLC + Random policy & 58.16 & 2.86\\ 
        MLC + Prediction policy & 63.85 & 2.95\\ 
        MLC + Sorted Pred policy & 48.36 & 2.92\\ \hline
        SC + Argmax policy & 62.87 & 2.94\\ \hline
        Clustering + TF-IDF & 61.76 & 2.83\\ \hline 
        Clustering + RoBERTa & 61.76 & 2.83\\ \hline 
        \hline
    \end{tabular}%
    }
    \caption{Performance of different routing models on the {\tt MMLU} data. {\sc Acc}: mean accuracy with {\sc maj@10} (\%), {\sc Lat}: LLM inference latency in seconds per query (10 generations for each query), {\sc mlc}: multi-label classifier, {\sc sc}: separate classifiers, and top-$k$: best $k$ performing models. All other notation is the same as for Table \ref{tab:routing_merge}.}
    \label{tab:det_routing_mmlu}
\end{table}

\newpage
\section{Related Work}
\label{sec:related}

\paragraph{Model Diversity} Several surveys~\cite[inter alia]{bommasani2021opportunities,minaee2024large} suggest that LLMs can develop emergent capabilities. Specifically, this suggests that models can show behavior and demonstrate skills beyond explicitly constructed ones. By virtue of differing training data, models may exhibit a wide variety of domain expertise. \citet{jiang-etal-2023-llm} demonstrates that no single open-source LLM outperforms other models across popular benchmarks. This further motivates the need to develop ensembling or routing methods aimed at improving the combined performance of a pool of LLMs with a diverse range of abilities.

\paragraph{Model Selection} A fundamental step in routing queries within an ensemble of models is to estimate the extent of overlap between the capabilities of the LLMs in the candidate pool with those deemed necessary to resolve an input query. Model selection in the context of LLM routing greatly differs from its traditional form in ML \cite{bishop_2006_pattern, raschka2020model}, wherein the training and test datasets are similar in distribution. Training data for LLMs include massive corpora spanning trillions of tokens with relatively straightforward learning objectives like next-token prediction \cite{brown2020language}. Test data, on the other hand, often involves highly structured tasks like reasoning and question answering \cite{hendryckstest2021, cobbe2021training, joshi-etal-2017-triviaqa}, summarization \cite{tam-etal-2023-evaluating}, and classification \cite{zhang2023sentiment}, which may not be very prevalent in corresponding training data. This makes gauging the pain points of resolving a complex query non-trivial. Furthermore, studies like ~\citet{rabinovich2023predicting} and~\citet{srivatsa2024naacl} suggest that certain aspects of the prompt phrasing, i.e., its length and readability, significantly impact LLMs' ability to tackle the underlying tasks.

\paragraph{LLM Ensembling} Previous attempts at ensembling and routing of LLMs aim to tackle one of two tasks: \textbf{(1)} Opting between LLM generations to select the best response. \citet{liu-liu-2021-simcls, ravaut-etal-2022-summareranker, jiang-etal-2023-llm} train models to rank or classify the most suitable response for a given query. However, this requires querying all LLMs in the model pool for each query during inference time. This can become computationally expensive with a large number of LLMs in the candidate pool. \textbf{(2)} Building routing networks ~\cite{rosenbaum2017routing} that utilize only a subset of parameters of a model or a subset of experts from a pool of candidate models. For example, \citet{jiang2024mixtral} employ a Mixture-of-Experts (MoE) ~\cite{jacobs1991adapt, collobert2002scaling, eigen2014learning} model with 8 experts, wherein only 2 experts are accessed at each model layer to produce the next token. This, however, requires pre-training the model weights, which incurs large computing and data costs. Alternatively, {\sc HybridLLM} ~\cite{ding2024hybrid}, \citet{shazeer2017outrageously}, and \citet{shnitzer2023large} train separate classifiers which select the best LLM(s) for each input query. 

This paper aims to create and study a sparse routing network for selecting the best LLM from a pool of more than two LLMs for each example. The routing network only needs to tune an extra Transformer-based classifier without needing to pre-train or fine-tune the LLMs. Furthermore, we also incorporate the former task by measuring the response quality (through accuracy) and determining if it can outperform the individual experts (LLMs) in the pool.

\end{document}